\documentclass{article}

\usepackage{arxiv}

\usepackage{amsmath}

\usepackage[utf8]{inputenc} % allow utf-8 input
\usepackage[T1]{fontenc}    % use 8-bit T1 fonts
\usepackage{hyperref}       % hyperlinks
\usepackage{url}            % simple URL typesetting
\usepackage{booktabs}       % professional-quality tables
\usepackage{amsfonts}       % blackboard math symbols
\usepackage{nicefrac}       % compact symbols for 1/2, etc.
\usepackage{microtype}      % microtypography
\usepackage{cleveref}       % smart cross-referencing
\usepackage{graphicx}
\usepackage{subcaption}
\usepackage[numbers]{natbib}
\usepackage{doi}

\usepackage{optidef}
\usepackage{bm}
\newcommand{\vv}[1]{\bm{#1}}
\newcommand{\diff}[2]{\frac{\mathrm{d}#1}{\mathrm{d}#2}}
\newcommand{\R}{\mathbb{R}}
\usepackage{physics}
\usepackage{float}
\usepackage{textcomp}

\title{Efficient Training of Physics-enhanced Neural ODEs via Direct Collocation and Nonlinear Programming\thanks{submitted to 16th International Modelica \& FMI Conference, May 2, 2025; accepted on June 23, 2025.}}

\date{August 1, 2025}

\usepackage{authblk}

\newbox{\orcid}\sbox{\orcid}{\includegraphics[scale=0.06]{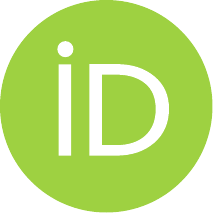}} 
\author[1]{
	\href{https://orcid.org/0009-0009-7517-4842}{\usebox{\orcid}\hspace{1mm}Linus Langenkamp}\thanks{corresponding author}}
\author[1]{
	\href{https://orcid.org/0009-0003-8902-9079}{\usebox{\orcid}\hspace{1mm}Philip Hannebohm}}
\author[1]{
	\href{https://orcid.org/0000-0002-4339-0438}{\usebox{\orcid}\hspace{1mm}Bernhard Bachmann}}
\affil[1]{Institute for Data Science Solutions, Bielefeld University of Applied Sciences and Arts, Germany}

\affil[ ]{\texttt{\{first.last\}@hsbi.de}}

\renewcommand{\shorttitle}{Efficient Training of PeN-ODEs via Direct Collocation and Nonlinear Programming}

\hypersetup{
	pdftitle={Efficient Training of Physics-enhanced Neural ODEs via Direct Collocation and Nonlinear Programming},
	pdfsubject={cs.LG, math.OC}
	pdfauthor={Linus Langenkamp, Philip Hannebohm, Bernhard Bachmann},
	pdfkeywords={Physics-enhanced Neural ODEs, Dynamic Optimization, Nonlinear Programming, Modelica, Neural ODEs, Universal Differential Equations},
}

\begin{document}
	\maketitle

	\begin{abstract}
		We propose a novel approach for training Physics-enhanced Neural ODEs (PeN-ODEs) by expressing the training process as a dynamic optimization problem. The full model, including neural components, is discretized using a high-order implicit Runge-Kutta method with flipped Legendre-Gauss-Radau points, resulting in a large-scale nonlinear program (NLP) efficiently solved by state-of-the-art NLP solvers such as Ipopt. This formulation enables simultaneous optimization of network parameters and state trajectories, addressing key limitations of ODE solver-based training in terms of stability, runtime, and accuracy. Extending on a recent direct collocation-based method for Neural ODEs, we generalize to PeN-ODEs, incorporate physical constraints, and present a custom, parallelized, open-source implementation. Benchmarks on a Quarter Vehicle Model and a Van-der-Pol oscillator demonstrate superior accuracy, speed, generalization with smaller networks compared to other training techniques. We also outline a planned integration into OpenModelica to enable accessible training of Neural DAEs.
	\end{abstract}
	
	\keywords{Physics-enhanced Neural ODEs \and Dynamic Optimization \and Nonlinear Programming \and Modelica \and Neural ODEs \and Universal Differential Equations}
	
	\section{Introduction}
	Growing access to real-world data and advances in computational modeling have opened new possibilities for combining measured data with physics-based models. Neural Ordinary Differential Equations (NODEs) \cite{Chen2018} represent a significant advancement in merging data-driven machine learning with physics-based modeling. By replacing the dynamics of an ODE with a neural network
	\begin{equation}
		\vv{\dot{x}}(t) = NN_{\vv{p}}(\vv{x}(t), \vv{u}(t), t),
	\end{equation}
	where $\vv{x}(t)$ are states, $\vv{u}(t)$ is a fixed input vector, and $\vv{p}$ are the neural network parameters, NODEs bridge the gap between traditional differential equations and modern deep learning. After obtaining a NODE and given an initial condition $\vv{x}(t_0) = \vv{x}_0$, the state trajectory is reconstructed by simulation with an arbitrary ODE solver
	\begin{equation}
		\vv{x}(t) := \mathrm{ODESolve}\left(NN_{\vv{p}}(\vv{x}(t), \vv{u}(t), t), \vv{x}_0\right). \label{eq:ivp}
	\end{equation}
	
	However, learning the full dynamics can be unstable, requires a lot of data, and can suffer from poor extrapolation \cite{Kamp2023}. As a result, hybrid modeling is an emerging field that combines the flexibility of neural networks with known physics and first principle models. Extensions like Universal Differential Equations (UDEs) \cite{Rackauckas2020UDE} or Physics-enhanced Neural ODEs (PeN-ODEs) \cite{Kamp2023, Sorourifar2023} generalize this paradigm, allowing domain-specific knowledge to be incorporated into the model while still learning observable but unresolved effects. These approaches have demonstrated great success in various fields, including vehicle dynamics \cite{Bruder2021, Thummerer2022FMU}, chemistry \cite{Thebelt2022}, climate modeling \cite{Ramadhan2023}, and process optimization \cite{Misener2023}.
	
	Training neural components typically involves simulating \eqref{eq:ivp} for some initial parameters $\vv{p}$ and then propagating sensitivities of the ODE solver backward in each iteration. Afterward, the parameters are updated via gradient descent. This process is computationally expensive and results in long training times \cite{Lehti2024, Roesch2021, Shapovalova2025}, as explicit integrators are low-order and unstable, requiring small step sizes, while  stable, implicit integrators involve solving nonlinear systems at each step, thus being computationally demanding.
	
	To address these enormous training times several alternative procedures have been proposed: in \cite{Roesch2021} a collocation technique is introduced, which approximates the right hand side (RHS) of an ODE from data. The NN is then trained on the approximations with standard training frameworks. Further, in \cite{Lehti2024} model order reduction is used to accurately simulate the dynamics in low-dimensional subspaces. Very recent work presented in \cite{Shapovalova2025} introduced global direct collocation with Chebyshev nodes, a method originating from dynamic optimization for optimal control and parameter optimization, for training Neural ODEs. The approach reduces the continuous training problem to a large finite dimensional nonlinear program (NLP) and shows fast and stable convergence, demonstrated on a typical problem, the Van-der-Pol oscillator.
	
	In Modelica-based workflows, the training of Neural ODEs is typically performed externally by exporting a Functional Mock-Up Unit (FMU), subsequently training it in Python or Julia using standard machine learning frameworks and ODE solvers, and finally re-importing the hybrid model. While this approach introduces external dependencies and additional transformation steps, the \textit{NeuralFMU} workflow \cite{Thummerer2022FMU} demonstrates a practical method for integrating hybrid models into real-world applications.
	
	Building on recent advances in direct collocation-based training of Neural ODEs \cite{Shapovalova2025}, we significantly extend the approach to PeN-ODEs. We formulate the training process as a dynamic optimization problem and discretize both neural and physical components using a stable, high-order implicit collocation scheme at flipped Legendre-Gauss-Radau (fLGR) points. This results in a large but structured NLP, allowing efficient, simultaneous optimization of states and parameters. Our custom, parallelized implementation leverages second-order information and the open-source NLP solver Ipopt \cite{Ipopt}. It is designed for a future integration into the open-source modeling and simulation environment OpenModelica \cite{Fritzson2020OM}, thus providing an accessible training environment independent of external tools.
	
	\section{Dynamic Optimization for NODEs} \label{sec:do_for_nodes}
	In this section, we introduce a general class of dynamic optimization problems (DOPs) and formulate training for both NODEs and PeN-ODEs as instances of this class. We then discuss the transcription of the continuous problem into a large-scale nonlinear optimization problem (NLP). Finally, necessary considerations and key challenges are presented.
	
	\subsection{Generic Problem Formulation}
	Consider the DOP
	\begin{mini!}
		{\vv{p}}{M(\vv{x}(t_0), \vv{x}(t_f), \vv{p}) + \int_{t_0}^{t_f} L(\vv{x}(t), \vv{p}, t) \, \mathrm{d}t}
		{}{\label{eq:opt_obj}}
		\addConstraint{\dot{\vv{x}}(t)}{= \vv{f}(\vv{x}(t), \vv{p}, t), \quad}{\qquad\qquad\forall t \in T \label{eq:dyn_eq}}
		\addConstraint{\vv{g}^L}{\leq \vv{g}(\vv{x}(t), \vv{p}, t)}{\leq \vv{g}^U, \quad}{\forall t \in T \label{eq:path_con}}
		\addConstraint{\vv{r}^L}{\leq \vv{r}(\vv{x}(t_0), \vv{x}(t_f), \vv{p})}{\leq \vv{r}^U \label{eq:boundary_con}}
	\end{mini!}
	for a fixed time horizon $T = [t_0, t_f]$ with time variable $t \in T$. The states of the system are given by $\vv{x}: T \to \R^{d_{\vv{x}}}$ and the goal is to find optimal time-invariant parameters $\vv{p} \in \R^{d_{\vv{p}}}$, such that the objective \eqref{eq:opt_obj} becomes minimal and the constraints \eqref{eq:dyn_eq}--\eqref{eq:boundary_con} are satisfied. These constraints are divided into the ODE \eqref{eq:dyn_eq} and \textit{path constraints} \eqref{eq:path_con}, which both must be satisfied at all times on time horizon $T$, as well as \textit{boundary constraints} \eqref{eq:boundary_con}, which must only hold at the initial and final time points $t_0, t_f$. The objective is composed of a \textit{Mayer} term $M$, that defines a cost at the boundary of $T$, and a \textit{Lagrange} term $L$, that penalizes an accumulated cost over the entire time horizon. To ensure compatibility with typical nonlinear optimizers, all model functions must be twice continuously differentiable. This includes neural networks, their activation functions, and error measures. For completeness, the bounds of the constraints are given as
	$\vv{g}^{L}, \vv{g}^{U} \in \left(\R \cup \{-\infty, \infty\}\right)^{d_{\vv{g}}}$ and  $\vv{r}^{L}, \vv{r}^{U} \in \left(\R \cup \{-\infty, \infty\}\right)^{d_{\vv{r}}}$.
	
	\subsection{Reformulation of PeN-ODE Training}
	\textit{PeN-ODEs} embed one or more NNs with parameters $\vv{p}$ into known, possibly equation-based, dynamics $\vv{\dot{x}}(t) = \vv{\hat{\phi}}(\vv{x}, \vv{u}, t)$. The resulting differential equation has the form
	\begin{equation}
		\vv{\dot{x}}(t) = \vv{\phi}(\vv{x}, \vv{u}, t, NN_{\vv{p}}(\vv{x}, \vv{u}, t)), \label{eq:PeN-ODE}
	\end{equation}
	where $\vv{u}: T \to \R^{d_{\vv{u}}}$ is a fixed input vector and $NN_{\vv{p}}$ are enhancing NNs. This formalism aims to enhance systems that already express dynamics based on first principles, by further incorporating data-driven observables in form of neural components. Clearly, these components need not be NNs in general, and can be any parameter dependent expression. With additional information about the problem, one could use a polynomial, rational function, sum of radial basis functions or Fourier series.
	
	The subsequent considerations also apply to the training of NODEs, where the goal is to learn the full dynamics without relying on a first principle model. This is evident from the fact that NODEs are a subclass of PeN-ODEs with
	\begin{equation}
		\vv{\dot{x}}(t)  =  NN_{\vv{p}}(\vv{x}, \vv{u}, t) = \vv{\phi}(\vv{x}, \vv{u}, t, NN_{\vv{p}}(\vv{x}, \vv{u}, t)).
	\end{equation}
	
	In this paper, we propose a formulation for training PeN-ODEs as a DOP \eqref{eq:opt_obj}--\eqref{eq:boundary_con}, using known data trajectories $\hat{\vv{q}}$ and the corresponding predicted quantity $\vv{q}$. The DOP takes the form
	\begin{mini!}
		{\vv{p}}{
			\int_{t_0}^{t_f} E\left(\vv{q}(\vv{x}, \vv{u}, t, NN_{\vv{p}}(\vv{x}, \vv{u}, t)), \hat{\vv{q}}(t)\right) \, \mathrm{d}t
		}{}{\label{eq:NN_obj}}
		\addConstraint{\dot{\vv{x}}(t)}{= \vv{\phi}\left(\vv{x}, \vv{u}, t, NN_{\vv{p}}(\vv{x}, \vv{u}, t)\right)}{\quad \forall t \in T \label{eq:NN_dyn}}{}
	\end{mini!}
	
	for some smooth error measure $E$, e.g. the squared $2$-norm $E(\vv{q}, \hat{\vv{q}})= \norm{\vv{q} - \hat{\vv{q}}}_2^2$. This formulation represents the minimal setup.
	
	By further incorporating the generic constraints \eqref{eq:path_con} and \eqref{eq:boundary_con}, it is possible to impose an initial or final condition on the states as well as enforce desired behavior. For example, consider a NN approximation of a force element $NN_F$, with no force acting in its resting position. Therefore, the NN should have a zero crossing, i.e. $NN_F(0) = 0$. This can be trivially formulated as a constraint, without introducing a penalty term that may distort the optimization as in standard unconstrained approaches like \cite{Kamp2023}. Thus, the optimizer can handle the constraint appropriately.
	
	As the loss in \eqref{eq:NN_obj} is a continuous-time integral, it enables an accurate and stable approximation using the same discretization employed for the system dynamics. In contrast to MSE loss as in \cite{Shapovalova2025}, which effectively corresponds to a first-order approximation of the integral, our formulation benefits from high-order quadrature, potentially preserving the accuracy of the underlying discretization.
	
	\subsection{Transcription with Direct Collocation}
	In the following, the general DOP \eqref{eq:opt_obj}--\eqref{eq:boundary_con} is reduced to a NLP using orthogonal direct collocation.
	Direct collocation approaches have proven to be highly efficient in solving DOPs and are implemented in a variety of free and commercial tools, such as PSOPT \cite{Becerra2010PSOPT}, CasADi \cite{Andersson2019Casadi} or GPOPS-II \cite{Patterson2014GPOPS}, as well as in Modelica-based environments like OpenModelica \cite{Ruge2014Collocation} or JModelica \cite{Magnusson2015JM}. While OpenModelica only supports optimal control problems, the other frameworks allow for simultaneous optimization of static parameters. Recent work in the field of NODEs \cite{Shapovalova2025} shows that learning the RHS of small differential equations can be performed stably and efficiently using global collocation with Chebyshev nodes.
	
	In direct collocation the states are approximated by piecewise polynomials, that satisfy the differential equation at so-called \emph{collocation nodes}, usually chosen as roots of certain orthogonal polynomials. If the problem is smooth and with increasing number of collocation nodes, these methods achieve \textit{spectral}, i.e. exponential, convergence to the exact solution. In this paper, the collocation nodes are chosen as the \textit{flipped Legendre-Gauss-Radau points (fLGR)} rescaled from $[-1, 1]$ to $[0,1]$. This rescaling is performed, so that the corresponding collocation method is equivalent to the Radau IIA Runge-Kutta method. Radau IIA has excellent properties, since it is $A$-, $B$- and $L$-stable and achieves order $2m-1$ for $m$ stages or collocation nodes. These nodes $c_j$ for $j=1, \ldots, m$ are given as the $m$ roots of the polynomial $(1-t) P_{m-1}^{(1,0)}(2t-1)$, where $P_{m-1}^{(1,0)}$ is the $(m-1)$-th Jacobi polynomial with $\alpha = 1$ and $\beta = 0$. A detailed explanation of the method's construction based on quadrature rules is given in \cite{Langenkamp2024}.
	
	First, we divide the time horizon $[t_0, t_f]$ into $n + 1$ intervals $[t_i, t_{i+1}]$ for $i=0,\ldots,n$ with length $\Delta t_i := t_{i+1} - t_{i}$. In each interval $[t_i, t_{i+1}]$ the collocation nodes $t_{ij} := t_i + c_j \Delta t_i$ for  $j=1, \ldots, m_i$ as well as the first grid point $t_{i0} := t_i + c_j \Delta t_i$ with $c_0 = 0$ are added.
	Since the last node $c_{m_i} = 1$ is contained in any Radau IIA scheme, the last grid point of interval $i-1$ exactly matches the first grid point of interval $i$, i.e. $t_{i-1,m_{i-1}} = t_{i0}$.
	Furthermore, the states are approximated as $\vv{x}(t_{ij}) \approx \vv{x}_{ij}$ and for each interval $i$ replaced by a Lagrange interpolating polynomial $\vv{x}_i(t) = \sum_{j=0}^{m_i} \vv{x}_{ij} \, l_j(t)$ of degree $m_i$, where
	\begin{equation}
		l_j(t) := \prod_{\substack{k = 0 \\ k \neq j}}^{m_i} \frac{t - t_{ik}}{t_{ij} - t_{ik}} \quad \forall j = 0, \ldots, m_i
	\end{equation}
	are the Lagrange basis polynomials. Note that the parameters $\vv{p}$ are time-invariant and thus, need not be discretized.
	Each $\vv{x}_i$ must satisfy the differential equation \eqref{eq:dyn_eq} at the collocation nodes $t_{ij}$ and also match the initial condition $\vv{x}_{i0}$, which is given from the previous interval $i-1$. By differentiating we get the collocated dynamics
	\begin{equation} \label{eq:discr_dynamics}
		\vv{0} =
		\begin{bmatrix}
			D^{(1)}_{10} I   & \ldots & D^{(1)}_{1m_i} I \\
			\vdots           & \ddots & \vdots \\
			D^{(1)}_{m_i0} I & \ldots & D^{(1)}_{m_im_i} I \\
		\end{bmatrix}
		\begin{bmatrix}
			\vv{x}_{i0} \\ \vdots \\ \vv{x}_{im_i}
		\end{bmatrix}
		-
		\Delta t_i
		\begin{bmatrix}
			\vv{f}_{i1} \\ \vdots \\ \vv{f}_{im_i}
		\end{bmatrix}
	\end{equation}
	with identity matrix  $I \in \R^{d_{\vv{x}} \times d_{\vv{x}}}$, entries of the first differentiation matrix $D^{(1)}_{jk} := \diff{\tilde{l}_k}{\tau}(c_j)$, where
	\begin{equation}
		\tilde{l}_k(\tau) := \prod_{\substack{r = 0 \\ r \neq k}}^{m_i} \frac{\tau - c_r}{c_k - c_r} \, \, \forall k = 0, \ldots, m_i,
	\end{equation}
	and the RHS of the ODE $\vv{f}_{ij} := \vv{f}(\vv{x}_{ij}, \vv{p}, t_{ij})$.
	The numerical values of $D^{(1)}_{jk}$ can be calculated very efficiently with formulas provided in \cite{Schneider1986}.
	
	Approximating $L$ is analogous to discretizing the differential equation. This is done by replacing the integral with a Radau quadrature rule of the form
	\begin{equation}
		\int_{t_0}^{t_f} L(\vv{x}(t), \vv{p}, t) \, \mathrm{d}t \approx \sum_{i=0}^{n} \Delta t_i  \sum_{j=1}^{m_i} b_j L(\vv{x}_{ij}, \vv{p}, t_{ij}),
	\end{equation}
	where the quadrature weights are given by
	\begin{equation}
		b_j = \int_{0}^{1}  \prod_{\substack{k = 1 \\ k \neq j}}^{m_i} \frac{\tau - c_k}{c_j - c_k} \, \mathrm{d}\tau \quad \forall j=1, \ldots, m_i. \label{eq:radau_weights}
	\end{equation}
	
	$M$ and the boundary constraints \eqref{eq:boundary_con} are approximated by replacing the values on the boundary with their discretized equivalents, i.e.
	$M(\vv{x}(t_0), \vv{x}(t_f), \vv{p}) \approx M(\vv{x}_{00}, \vv{x}_{nm_n}, \vv{p})$ and $\vv{r}(\vv{x}(t_0), \vv{x}(t_f), \vv{p}) \approx \vv{r}(\vv{x}_{00}, \vv{x}_{nm_n}, \vv{p})$, while the path constraints \eqref{eq:path_con} are evaluated at all nodes, i.e. $\vv{g}(\vv{x}(t_{ij}), \vv{p}, t_{ij}) \approx \vv{g}(\vv{x}_{ij}, \vv{p}, t_{ij})$.
	
	\subsection{Training with Nonlinear Programming}
	By flattening the collocated dynamics \eqref{eq:discr_dynamics}, we obtain the discretized DOP \eqref{eq:opt_obj}--\eqref{eq:boundary_con} of the form
	\begin{mini!}
		{\vv{x}_{ij}, \vv{p}}{
			M(\vv{x}_{00}, \vv{x}_{nm_n}, \vv{p}) + \sum_{i=0}^{n} \Delta t_i \sum_{j=1}^{m_i} b_j L(\vv{x}_{ij}, \vv{p}, t_{ij})
		}{}{\label{eq:d_obj}}
		\addConstraint{\vv{0}}{= \sum_{k=0}^{m_i} D^{(1)}_{jk} \vv{x}_{ik} - \Delta t_i \vv{f}(\vv{x}_{ij}, \vv{p}, t_{ij})}{}{\qquad \forall\, i, \forall j \geq 1 \label{eq:d_dyn}}{}
		\addConstraint{\vv{g}^L}{\leq \vv{g}(\vv{x}_{ij}, \vv{p}, t_{ij})}{\leq \vv{g}^U}{\qquad\qquad\quad\,\,\,\,\, \forall\, i, \forall j \geq 1 }{\label{eq:d_path}}
		\addConstraint{\vv{r}^L}{\leq \vv{r}(\vv{x}_{00}, \vv{x}_{nm_n}, \vv{p})}{\leq \vv{r}^U}{\label{eq:d_boundary}}
	\end{mini!}
	
	This large-scale NLP \eqref{eq:d_obj}--\eqref{eq:d_boundary} can be implemented and solved efficiently in nonlinear optimizers such as Ipopt \cite{Ipopt}, SNOPT \cite{SNOPT1, SNOPT2} or KNITRO \cite{Knitro}. These NLP solvers exploit the sparsity of the problem as well as the first and second order derivatives of the constraint vector and objective function to converge quickly to a suitable local optimum.
	
	The open-source interior-point method {Ipopt} requires the already mentioned first derivatives and, in addition, the Hessian of the augmented Lagrangian at every iteration.
	Since the derivatives only need to be evaluated at the collocation nodes, there is no need to propagate them as in traditional ODE solver-based training. Furthermore, training with ODE solvers usually limits itself to first order derivatives and therefore, does not utilize higher order information as in the proposed approach.
	
	The resulting so-called \textit{primal-dual} system is then solved using a linear solver for symmetric indefinite systems, such as the open-source solver MUMPS \cite{mumps} or a proprietary solver from the HSL suite \cite{hsl}, after which an optimization step is performed. This step updates all variables $\vv{x}_{ij}, \vv{p}$ simultaneously, allowing for direct observation and adjustment of intermediate values. In contrast, ODE solver training only captures the final result after integrating the dynamics over time, without the ability to directly influence intermediate states during the optimization process.
	Because this linear system must be solved in every iteration anyway, high order, stable, implicit Runge-Kutta collocation methods, e.g. Radau IIA, can be embedded with only limited overhead. As a result, the NLP formulation overcomes key limitations of explicit ODE solvers in terms of order, stability, and allowable step size. Moreover, since the solver performs primal and dual updates, the solution does not need to remain feasible during the optimization, in contrast to ODE solver approaches where the dynamics \eqref{eq:dyn_eq} are enforced at all times through forward simulation.
	This results in both advantages and disadvantages: On the one hand, it enables more flexible and aggressive updates, potentially accelerating convergence. On the other hand, it may lead to intermediate solutions that temporarily violate physical consistency or produce invalid function evaluations, which require careful handling.
	
	For a comprehensive overview of nonlinear programming, interior-point methods, and their application to collocation-based dynamic optimization, we refer interested readers to \cite{Biegler2010NLP} and the Ipopt implementation paper \cite{Ipopt}.
	
	\subsection{Challenges and Practical Aspects}
	We identify four main challenges in training PeN-ODEs using the proposed direct collocation and nonlinear programming approach. These challenges are closely related to those encountered in conventional PeN-ODE or general NN training.
	
	\subsubsection{Grid Selection} \label{sec:grid}
	The choice of time grid $\{t_0,\dots,t_{n+1}\}$ and the number of collocation nodes per interval $m_i$ are crucial for both the accuracy and efficiency of the training process. In practice, the grid can either be chosen equidistant or tailored to the specific problem. While equidistant grids are straightforward to implement and often sufficient for well-behaved systems, non-equidistant grids may reduce computational costs while capturing the dynamics more efficiently. Placing more intervals with low degree collocation polynomials in regions of rapid state change can improve approximation quality without unnecessarily increasing the problem size. Similarly, in well-behaved regions, it is feasible to perform larger steps with more collocation nodes. Because the collocation scheme and grid are embedded into the NLP, these must be given \textit{a-priori}. This leaves room for future developments of adaptive mesh refinement methods with effective mesh size reduction, which have already shown great success for optimal control problems \cite{Zhao2018, Liu2015}.
	
	\subsubsection{Initial Guesses} \label{sec:initial_guess}
	Due to the size and possible nonlinearity of the resulting NLP, the choice of initial guesses has a strong influence on convergence behavior. Unlike classical NN training, where poor initialization primarily affects convergence speed, the constrained nature of the transcribed dynamic optimization problem can lead to poor local optima or even solver failure. It is therefore of high importance to perform informed initializations for the states $\vv{x}_{ij}$ and, if possible, for the NN parameters $\vv{p}$.
	
	One practical approach to obtain the required parameter guesses is to first train the network on a small, representative subset of the full dataset using constant initial values for both the states and parameters. The optimized parameters resulting from this reduced problem then serve as informed initial guesses for the full training problem. Consequently, the states are obtained by simulation, i.e. $\vv{x}(t) := \mathrm{ODESolve}\left(\vv{\phi}(\vv{x}, \vv{u}, t, NN_{\vv{p}}(\vv{x}, \vv{u}, t), \vv{x}_0)\right)$ and $\vv{x}_{ij} := \vv{x}(t_{ij})$ for a given initial condition $\vv{x}(t_0) = \vv{x}_0$. By construction, the collocated dynamic constraints \eqref{eq:d_dyn} are satisfied, leading to improved convergence and stability in the full NLP.
	
	Clearly, this strategy does not work in general. However, in simple cases where the model can be decomposed and the NN's input-output behavior is observable, e.g. if model components should be replaced by a neural surrogate, the NN can be pre-trained using standard gradient descent. This yields reasonable initial guesses for the parameters and states by simulation, which can then be integrated into the constrained optimization problem. Another pre-training strategy could be the \textit{collocation technique} proposed in \cite{Roesch2021}. Still, developing general, effective strategies to obtain reasonable initial guesses is one of the key challenges and limiting factors we identify for the general application of this approach.
	
	\subsubsection{Batch-wise Training} \label{sec:batch}
	Standard ML frameworks employ batch learning to efficiently split up data. This is not as straightforward when training with the approach described here. One might assume that the entire dataset must be included in a single discretized DOP. However, recent work \cite{Shapovalova2025} demonstrates that batch-wise training is possible and promises significant potential. The Alternating Direction Method of Multipliers (ADMM) \cite{ADMM} allows decomposing the optimization problem into smaller subproblems that can be trained independently, while enforcing consensus between them. This allows for memory-efficient training and opens up the possibility of handling larger models or learning from multiple data trajectories simultaneously.
	
	\subsubsection{Training of Larger Networks} \label{sec:largernn}
	While computing Hessians of NNs is generally expensive, it is tractable for small networks. To reduce computational effort for larger networks, it might be reasonable to use partial Quasi-Newton approximations such as SR1, BFGS or DFP
	to approximate the dense parts of the augmented Lagrangian Hessian $H$, e.g. the blocks $H_{\vv{x}_{ij},\vv{p}}$ and $H_{\vv{p}\vv{p}}$, or solely $H_{\vv{p}\vv{p}}$.
	These blocks, which contain derivatives with respect to the NN parameters, are computationally expensive, while the block $H_{\vv{x}_{ij}, \vv{x}_{ij}}$, which contains second derivatives with respect to the collocated states, is extremely sparse and comparably cheap. A Quasi-Newton approximation of the block $H_{\vv{x}_{ij}, \vv{x}_{ij}}$ is therefore disadvantageous. The sparsity can be exploited by computing this block analytically and using it directly in the Quasi-Newton update. An implementation of this partial update using the SR1 Quasi-Newton method is straightforward. Instead of one expensive symmetric rank-one update for the entire Hessian $H$, one cheap symmetric rank-one update for $H_{\vv{p}\vv{p}}$ and one general rank-one update for $H_{\vv{x}_{ij},\vv{p}}$ are needed.
	
	This procedure significantly reduces the cost of the Hessian, while still providing fairly detailed derivative information. SR1 is particularly advantageous here, as it can represent indefinite Hessians, which is favorable when dealing with highly nonlinear functions.
	
	Since our current examples perform well with small NNs, we do not explore larger networks in this paper. However, we anticipate that such Quasi-Newton strategies will be necessary in future work with larger networks. Very recent work \cite{Lueg2025newNDAE} independently expresses similar ideas, highlighting the potential of the approach.
	
	\section{Implementation} \label{sec:implementation}
	The generic DOP \eqref{eq:opt_obj}--\eqref{eq:boundary_con} and its corresponding NLP formulation \eqref{eq:d_obj}--\eqref{eq:d_boundary} are implemented in the custom open-source framework {GDOPT} \cite{Langenkamp2024}, which is publicly available.\footnote{\url{https://github.com/linuslangenkamp/GDOPT}} For neural network training the code has been extended, including predefined parametric blocks such as neural networks, support for data trajectories and parallelized optimizations. This extended, experimental version is also publicly available.\footnote{\url{https://github.com/linuslangenkamp/GDOPT_DEV}}
	
	\subsection{GDOPT}
	GDOPT (General Dynamic Optimizer) consists of two main components: an expressive Python-based package \textit{gdopt} and an efficient C++ library \textit{libgdopt}. The Python interface provides an user-friendly modeling environment and performs symbolic differentiation and code generation. Symbolic expressions are optimized using common subexpression elimination via SymEngine\footnote{\url{https://github.com/symengine/symengine}}, and the resulting expressions together with first and second derivatives are translated into efficient C++ callback functions for runtime evaluation. In the present implementation, all expressions are flattened, resulting in large code, especially for the Hessian. Note that keeping NNs vectorized and employing symbolic differentiation rules with predefined NN functions offers significant advantages.
	
	The library \textit{libgdopt} implements a generalized version of the NLP \eqref{eq:d_obj}--\eqref{eq:d_boundary} using Radau IIA collocation schemes, while also supporting optimal control problems. It is interfaced with Ipopt to solve the resulting nonlinear programs. Both symbolic Jacobian and Hessian rely on exact sparsity patterns discovered in the Python interface.
	Additional functionality includes support for nominal values, initial guesses, runtime parameters, mesh refinements, plotting utilities, and special functions. A detailed overview of features and modeling is provided in the GDOPT User’s Guide\footnote{\url{https://github.com/linuslangenkamp/GDOPT/blob/master/usersguide/usersguide.pdf}}.
	
	Nevertheless, GDOPT lacks important capabilities that established modeling languages and tools offer, such as object-oriented, component-based modeling and support for differential-algebraic equation \cite{Akesson2012DAE} systems. It is possible to model DAEs by introducing control variables for algebraic variables. However, this approach increases the workload and size of the NLP and may lead to instabilities.
	
	\subsection{Parallel Callback Evaluations} \label{sec:para}
	Since in every optimization step, the function evaluations as well as first and second derivatives of all NLP components \eqref{eq:d_obj}--\eqref{eq:d_boundary} must be provided to Ipopt, an efficient callback evaluation is crucial to accelerate the training. Note that these callbacks themselves consist of the continuous functions evaluated at all collocation nodes. For simplicity we write $\vv{z}_{ij} := [\vv{x}_{ij}, \vv{p}]^T$ for the variables at a given collocation node and, in addition,
	\begin{equation}
		\vv{\psi}(\vv{x},\vv{p}, t) := [L(\cdot) , \vv{f}(\cdot) , \vv{g}(\cdot) ]^T
	\end{equation}
	for the vector of functions that are evaluated at all nodes. Clearly, the required callbacks
	\begin{equation*}
		\vv{\psi}\big|_{z_{ij}} \quad \nabla \vv{\psi} \big|_{z_{ij}} \quad \nabla^2 \vv{\psi} \big|_{z_{ij}} \quad \forall i \,\, \forall j\geq 1
	\end{equation*}
	are independent with respect to the given collocation nodes $t_{ij}$ and thus, allow for a straightforward parallelization. In the extension of GDOPT, we use OpenMP \cite{OpenMP} to parallelize the callback evaluations required by Ipopt. Depending on the specific callback, i.e. objective function, constraint violation, gradient, Jacobian, or Hessian of the augmented Lagrangian, a separate \texttt{omp parallel for} loop is used to evaluate the corresponding components at all collocation nodes.
	This design results in a significant reduction in computation time, especially for the comparably expensive dense derivatives of neural components.
	
	\section{Performance} \label{sec:results}
	In order to test the proposed training method and parallel implementation, two example problems are considered. The first example is the Quarter Vehicle Model (QVM) from \cite{Kamp2023}, where an equation-based model is enhanced with small neural components, such that physical behavior is represented more accurately. The second example is a standard NODE, where the dynamics of a Van-der-Pol oscillator \cite{Roesch2021} are learned purely from data. In both cases, the experimental setup closely follows the configurations used in the respective paper. Training is performed on a laptop running Ubuntu 24.04.2 with Intel Core i7-12800H ($20$ threads), 32 GB RAM and using GCC v13.3.0 with flags \texttt{-O3 -ffast-math} for compilation, while MUMPS \cite{mumps} is used to solve linear systems arising from Ipopt \cite{Ipopt}. All dependencies are free to use and open-source.
	
	\subsection{Quarter Vehicle Model}
\begin{figure}[h]
		\centering
		\includegraphics[width=.35\textwidth, angle=-90]{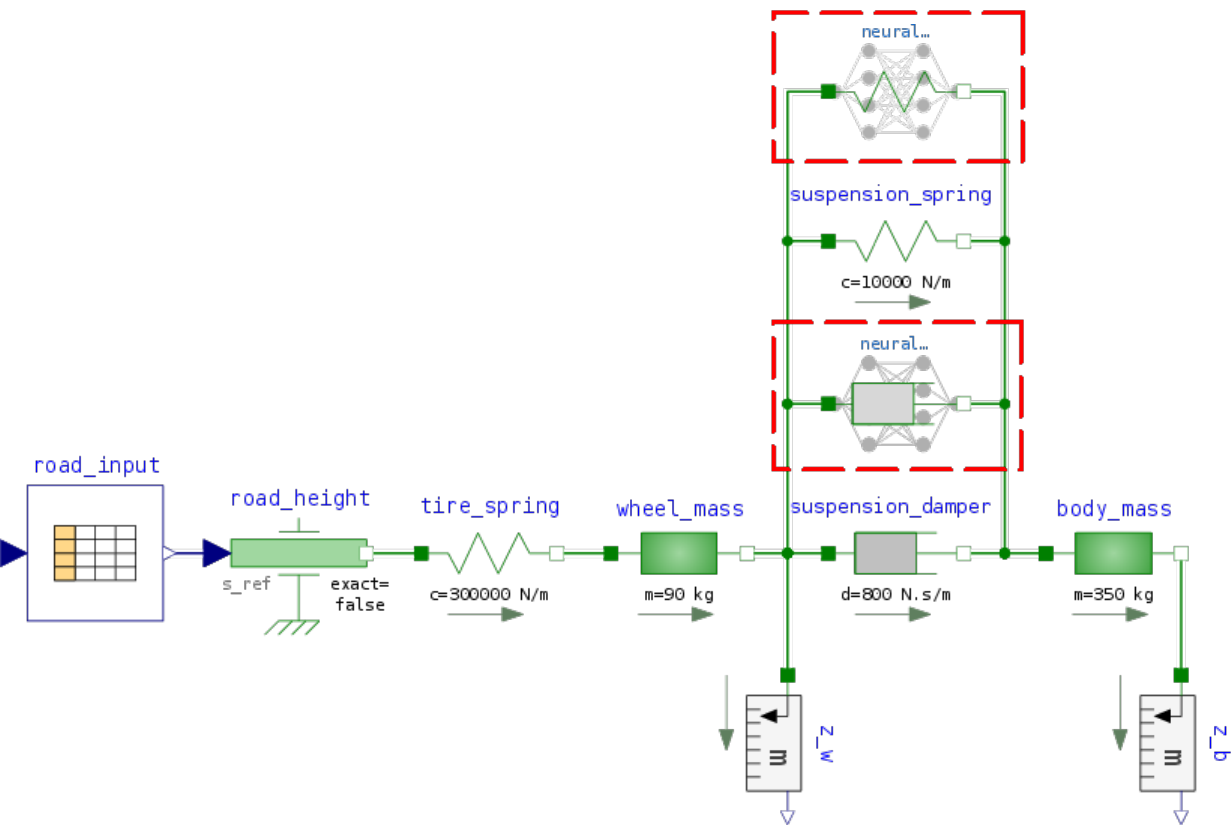}
		\caption{Modelica Models of the Linear (without boxes) and Neural (with boxes) QVM.
			Modified from T. Kamp.}
		\label{fig:qvm_mo}
	\end{figure}

	We follow the presentation in \cite{Kamp2023} for an overview of the model and the data generation process.
	The Quarter Vehicle Model (QVM) captures the vertical dynamics of a road vehicle by modeling one wheel and the corresponding quarter of the vehicle body. It consists of two masses connected by spring-damper elements representing the suspension and tire dynamics.
	The linear base model is described by the differential equations  $\dot{z}_r = u$, $\dot{z}_b = v_b$, $\dot{z}_w = v_w$, and
	\begin{subequations}
		\begin{align}
			\dot{v}_b &:= a_b = m_{b}^{-1} \left( c_s \Delta z_s + d_s \Delta v_s \right) \label{eq:qvm_1}\\
			\dot{v}_w &:= a_w = m_{w}^{-1} \left( c_t \Delta z_t + d_t \Delta v_t - c_s \Delta z_s - d_s \Delta v_s \right) \label{eq:qvm_2}
		\end{align}
	\end{subequations}
	where $\Delta z_s = z_w - z_b$, $\Delta v_s = v_w - v_b$, $\Delta z_t = z_r - z_w$, $\Delta v_t = u - v_w$. In addition, $m_w$ is the mass of the wheel, $m_b$ is mass of the quarter body, and $c_s$ and $d_s$ are the coefficients of the linear spring-damper pair between these masses, modeling the suspension. Furthermore, $c_t$ and $d_t$ define an additional linear spring-damper pair between the tire and ground. The state vector $\vv{x} = [z_b, z_w, v_b, v_w, z_r]^T$ contains the positions of the body $z_b$ and wheel $z_w$, their velocities $v_b$ and $v_w$, and the road height $z_r$. The differential road height $\dot{z}_r = u$ is given as an input and the observable outputs $y = [a_w, a_b]$ measure the wheel and body accelerations. A corresponding Modelica model of the linear QVM is depicted in Figure \ref{fig:qvm_mo}. \cite{Kamp2023}
	
	\subsubsection{Data Generation}
	The linear model is extended by introducing two additional nonlinear forces between both masses, a translational friction force $F_{fr}$ and a progressive spring characteristic $F_{pr}$. These nonlinear components are introduced only for data generation, creating a more complicated, nonlinear model whose behavior deviates from the known base dynamics. Therefore, the differential equations \eqref{eq:qvm_1} and \eqref{eq:qvm_2} become
	\begin{subequations}
		\begin{align}
			\dot{v}_b &= m_{b}^{-1} \left( c_s \Delta z_s + d_s \Delta v_s + F_{pr}(\Delta z_s) + F_{fr}(\Delta v_s) \right) \\
			\dot{v}_w &=  m_{w}^{-1} ( c_t \Delta z_t + d_t \Delta v_t - c_s \Delta z_s - d_s \Delta v_s \nonumber \\ &\quad\quad\quad - F_{pr}(\Delta z_s) - F_{fr}(\Delta v_s) )
		\end{align}
	\end{subequations}
	To generate data, a simulation of the nonlinear model for an imitation of a realistic, rough road (ISO$8608$, Type D \cite{Mucka2018}) is performed. The observables $a_b$ and $a_w$ as well as the states are disturbed by random Gaussian noise and sampled with $1000$ Hz for a $42$ s trajectory as in \cite{Kamp2023}.
	
	\subsubsection{Training Setup}
	The nonlinear components $F_{fr}(\Delta v_s)$ and $F_{pr}(\Delta z_s)$ introduced for data generation are replaced by two neural network surrogates $F_{fr}^{NN}(\Delta v_s)$ and $F_{pr}^{NN}(\Delta z_s)$, illustrated in Figure \ref{fig:qvm_mo}. The goal is to find suitable replacements that minimize the mismatch between the simulated, disturbed outputs $\hat{a}_b$ and $\hat{a}_w$ of the nonlinear model and the observed data during the optimization. As discussed before, we write this objective as an integral over the entire time horizon, i.e.
	\begin{equation}
		\min \int_{t_0}^{t_f} \left(\frac{\hat{a}_b - a_b}{\sigma_{\hat{a}_b}}\right)^2 + \left(\frac{\hat{a}_w - a_w}{\sigma_{\hat{a}_w}}\right)^2 \, \dd t,
	\end{equation}
	where $\sigma_{\hat{a}_b}$ and $\sigma_{\hat{a}_w}$ are corresponding standard deviations of the data, ensuring that both accelerations contribute equally to the objective.
	Furthermore, it is known that both force elements have a zero crossing and therefore, the additional constraints
	\begin{equation}
		F_{fr}^{NN}(0) = 0 \quad \text{and} \quad F_{pr}^{NN}(0) = 0
	\end{equation}
	are simply added to the optimization problem.
	
	We employ three different training strategies. At first, both feedforward neural networks are trained directly on the full trajectory using randomly initialized parameters, while the initial state guesses are obtained from a simulation of the linear model (I). Each network has the structure $1 \cross 5\to 5 \cross 5 \to 5 \cross 1$ and therefore, both nets contain just $92$ parameters in total. We use the smooth squareplus activation function
	\begin{equation}
		\mathrm{squareplus}(x) := \frac{x + \sqrt{1 + x^2}}{2}
	\end{equation}
	to ensure a twice continuously differentiable NN as required for Ipopt. In the second strategy (II), described in Section \ref{sec:initial_guess}, first an acceleration scheme is employed, where the same networks are trained on a short segment one eighth of the entire trajectory. After that, a simulation of the neural QVM with the obtained parameters is performed. The resulting states are used as initial guesses in the subsequent optimization with full data. To show that the surrogates need not be NNs and training can be performed efficiently with other parameter-dependent expressions, the third strategy (III) uses rational functions to model the unknown behavior. For instance, $F_{fr}$ is replaced by
	\begin{equation}
		F_{fr}^{RC}(\Delta v_s) := \frac{\sum_{k=0}^{N} \omega_k T_k\left(\Delta v_s\right)}{\sum_{k=0}^{D} \theta_k T_k\left(\Delta v_s\right)},
	\end{equation}
	where $T_k$ is the $k$-th Chebyshev polynomial, $\omega_k$ and $\theta_k$ are parameters to be optimized, and $N$ and $D$ are the numerator and denominator degrees. For both rational functions we choose $N = D = 7$, resulting in a total of merely $32$ learnable parameters.
	
	Since the QVM contains very fast dynamics due to high-frequency excitations, in all cases the time horizon is divided into a tightly spaced, equidistant grid of $2500$ intervals and using a constant $5$-step Radau IIA collocation scheme of order $9$. This leads to a total of $12500$ collocation nodes and more than $2.7\times10^{6}$ nonzeros in the Jacobian and roughly $4.73\times10^{6}$ nonzeros in the Hessian of the large-scale NLP.
	\subsubsection{Results}
\begin{figure}[h]
		\centering
		\begin{minipage}{0.4\textwidth}
			\centering
			\vfill
			\begin{table}[H]
				\centering
				\vspace{42pt}
				\begin{tabular}{@{}lrrr@{}}
					\toprule
					\textbf{Strategy} & \textbf{Total} & \textbf{Ipopt} & \textbf{Callbacks} \\
					\midrule
					(I) & 394.43 & 284.18 & 110.25 \\
					\midrule
					(II) - initial &  26.75 & 15.08 & 11.67 \\
					(II) - final & 173.06 & 124.65 & 48.41 \\
					(II) & 199.81 & 139.73 & 60.07 \\
					\midrule
					(III) & 34.96 & 27.10 & 7.85 \\
					\bottomrule
				\end{tabular}
				\vspace{34pt}
				\caption{Training Times in Seconds}
				\label{tab:timings_1}
			\end{table}
		\end{minipage}
		\hfill
		\begin{minipage}{0.5\textwidth}
			\centering
			\includegraphics[width=1\textwidth]{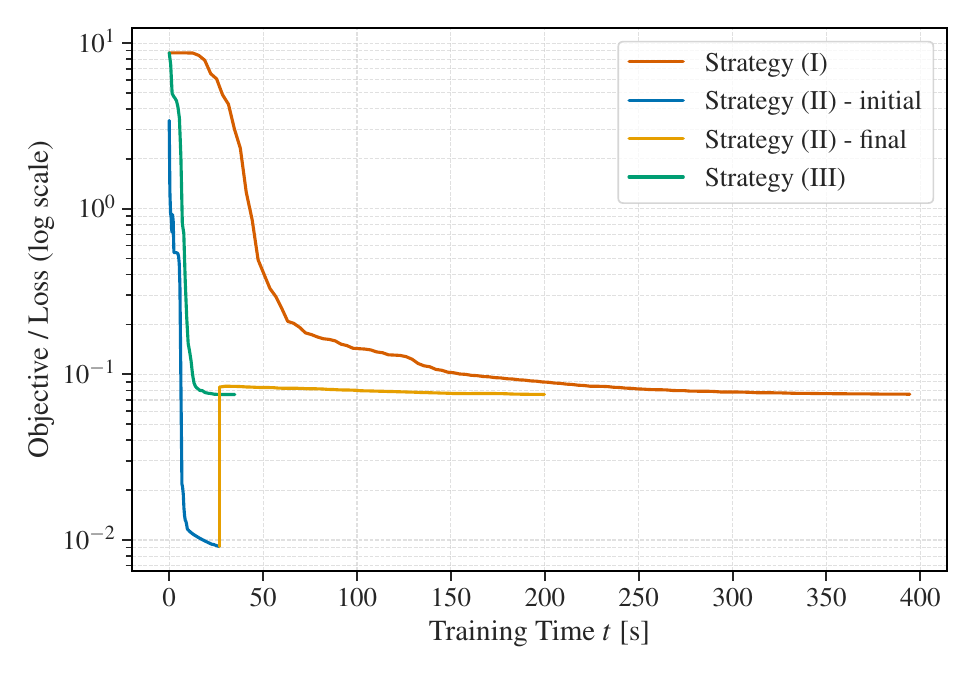}
			\vspace{-15pt}
			\caption{Objective History with Respect to Training Time}
			\label{fig:losses}
		\end{minipage}
	\end{figure}

	Table \ref{tab:timings_1} presents the training times for all strategies. Each optimization was run for a maximum of $150$ NLP iterations and was automatically terminated early if no further significant improvement in objective could be achieved. We want to stress that all trainings, performed on a laptop, are executed in under $7$ minutes, compared to $4.5$ hours for the fastest optimization in \cite{Kamp2023} using ODE solver-based training. Clearly, this is also due to the fact that smaller neural components are used. Furthermore, Figure \ref{fig:losses} depicts the objective value with respect to training time, where both the first and second optimizations of (II) are concatenated.
	
\begin{figure}[H]
		\centering
		\begin{minipage}{0.48\textwidth}
			\centering
			\includegraphics[width=\linewidth]{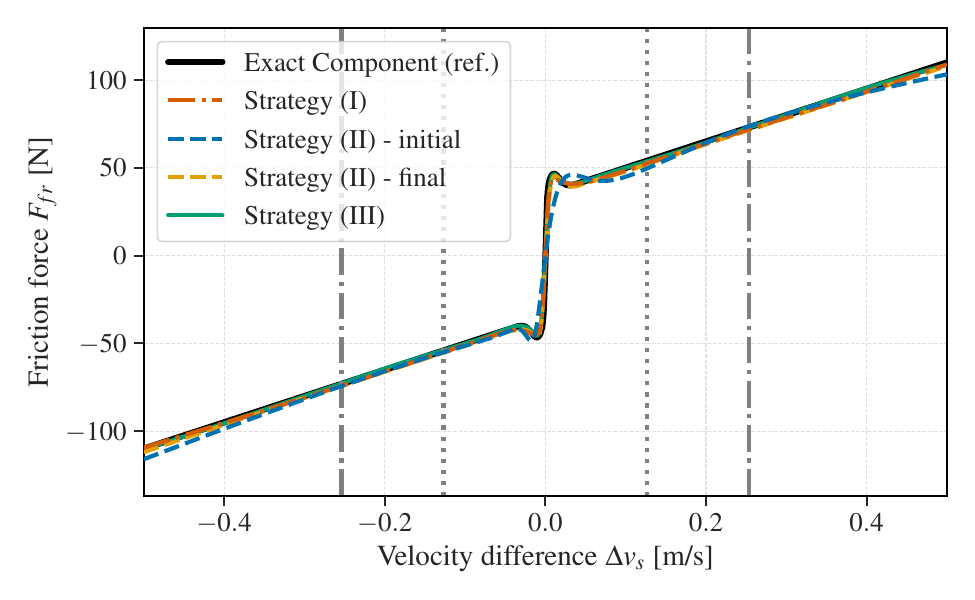}
			\vspace{-15pt}
			\caption{Neural and Reference Damper Characteristics}
			\label{fig:damper}
		\end{minipage}
		\hfill
		\begin{minipage}{0.48\textwidth}
			\centering
			\includegraphics[width=\linewidth]{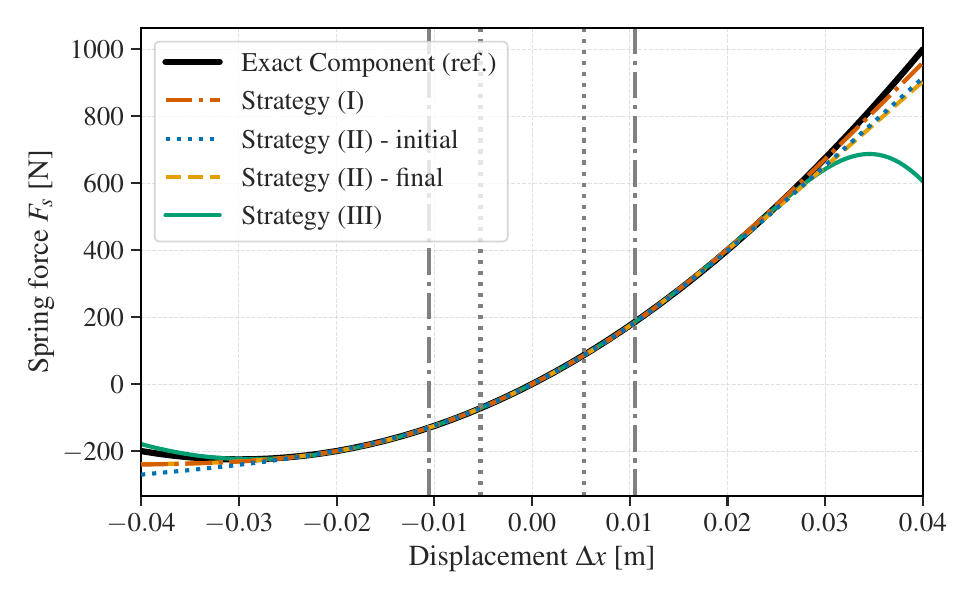}
			\vspace{-15pt}
			\caption{Neural and Reference Spring Characteristics}
			\label{fig:spring}
		\end{minipage}
	\end{figure}

	Even though a poor initialization has been used and thus the optimization required more time, the naive strategy (I) converged stably to a suitable optimum with a similar objective as strategies (II) and (III). For this example, the acceleration scheme (II) proves effective, since the initial optimization for a shorter data trajectory takes just $26.75$ s in total, and the subsequent initial guesses for states and parameters become very good approximations of the real solution. This can be observed in Figure \ref{fig:damper} and Figure \ref{fig:spring}, where the resulting neural surrogates are depicted. By performing the second optimization, (II) effectively halves the time required by strategy (I), i.e. less than $3.5$ minutes, and furthermore results in indistinguishable neural components.
	
\begin{figure}[h!]
		\centering
		\includegraphics[width=0.6\textwidth]{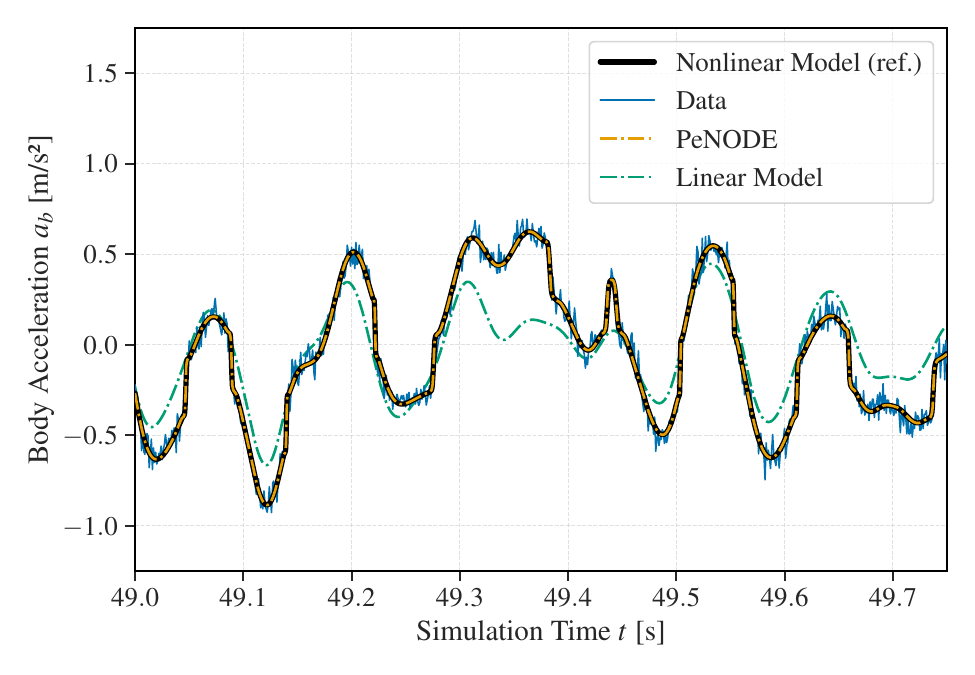}
		\caption{Body Accelerations $a_b$ for Simulations of PeN-ODE and Standard Models on a Type C Road \cite{Mucka2018}}
		\label{fig:simulation}
	\end{figure}

	Moreover, as seen in Figure \ref{fig:damper} and Figure \ref{fig:spring}, the resulting very small NNs match the reference in almost perfect accordance, correctly representing highly nonlinear parts of the damper. These also generalize in a very natural way, as can be seen from the behavior outside the vertical dotted lines, which represent the first and second standard deviations of the inputs to the nets. Performing simulations on a new, unknown input road shows that the obtained PeN-ODE from (II) matches the nonlinear reference model perfectly, as illustrated in Figure \ref{fig:simulation}.
	Note that, our characteristics of the damper and spring serve as even better surrogates than those reported in \cite{Kamp2023}, despite using significantly smaller networks and requiring considerably shorter training times. While \cite{Kamp2023} relied on larger models with longer ODE solver-based training, our approach yields more accurate and better-generalizing results.
	
	By having principal knowledge of the underlying characteristics shown in Figure \ref{fig:damper} and Figure \ref{fig:spring}, it is possible to model observed behavior with minimal parameters. Since the number of Hessian nonzeros grows quadratically with the number of parameters, such knowledge of the procedure can greatly benefit both training time and surrogate quality. Therefore, consider simple rational functions as an educated guess for both missing components. This optimization, with $32$ instead of $92$ free parameters, is performed in under $35$ seconds without any acceleration strategy. Moreover, the obtained surrogates are of mostly equal quality to the NN components from (II), although Figure \ref{fig:spring} shows that the rational function does not generalize as well. Nevertheless, these results demonstrate that unknown behavior may be expressed with fewer parameters and yield equivalent quality.
	
	\subsubsection{Parallel Implementation}
	Finally, it is stressed that the parallel implementation, described in Section \ref{sec:para}, leads to $5.44$ times less time taken in the generated function callbacks. This yields a total training time that is more than halved and clearly shows that the implementation efficiently exploits the independence of collocation nodes.
	
	\begin{table}[h]
		\centering
		\begin{tabular}{@{}crrr@{}}
			\toprule
			\textbf{Method} & \textbf{Total} & \textbf{Ipopt} & \textbf{Callbacks} \\
			\midrule
			GDOPT (default)  & 385.90 & 122.52 & 263.38 \\
			GDOPT (parallel)  & 173.06  & 124.65 & 48.41 \\
			\bottomrule
		\end{tabular}
		\vspace{10pt}
		\caption{Comparison of Parallel and Sequential Optimization Times in Seconds of (II) - final}
		\label{tab:timings}
	\end{table}
	
	\subsection{Van-der-Pol Oscillator}
	To illustrate the ability of learning a full NODE, we follow an example from \cite{Roesch2021}, where a different kind of {collocation method} was proposed. This method approximates the RHS of the ODE with data, thus enabling faster, unconstrained training without the need for ODE solvers. Consider the Van-der-Pol (VdP) oscillator
	\begin{subequations}
		\begin{align}
			\dot{x} &= y \\
			\dot{y} &= \mu  y \left(1 - x^2\right) - x
		\end{align}
	\end{subequations}
	with $\mu = 1$ and initial conditions $x(t_0) = 2$ and $y(t_0) = 0$. Data generation is performed by simulating the dynamics with OpenModelica \cite{Fritzson2020OM} on an equidistant grid with $200$ intervals and artificially perturbing the observed states by additive Gaussian noise. To test sensitivities of the approach, we use 3 different levels of noise $\mathcal{N}(0, \sigma)$ to disturb the observable states, i.e. no noise ($\sigma=0$), low noise ($\sigma=0.1$) and high noise ($\sigma=0.5$).
	
	The continuous DOP has the form
	\begin{mini!}
		{\vv{p}_x, \vv{p}_y}{ 
			\int_{t_0}^{t_f} (\hat{x}_{\sigma} - x)^2 + (\hat{y}_{\sigma} - y)^2 \, \mathrm{d}t + \lambda \norm{\vv{p}}_2^2
		}{\label{eq:vdpObj}}{}
		\addConstraint{\dot{x}}{= NN_{\vv{p}_x}^{x}(x, y)}{}
		\addConstraint{\dot{y}}{= NN_{\vv{p}_y}^{y}(x, y)}{}
		\addConstraint{x(t_0)}{= 2}{}
		\addConstraint{y(t_0)}{= 0}{}
	\end{mini!}
	where $\hat{x}_{\sigma}, \hat{y}_{\sigma}$ is the disturbed state data, $NN_{\vv{p}_x}^{x}(x, y)$, $NN_{\vv{p}_y}^{y}(x, y)$ are neural networks of the architecture $2 \cross 5\to 5 \cross 5 \to 5 \cross 1$ with $\mathrm{sigmoid}$ activation function, $\vv{p} = [\vv{p}_x, \vv{p}_y]^T$ are $102$ learnable parameters, while $\lambda > 0$ is a regularization factor to enhance stability.
	
	As in \cite{Roesch2021}, the training is performed on a $7$ second time horizon, thus including a little over one period of the oscillator. Furthermore, $500$ intervals and the $5$-step Radau IIA method of order $9$ are used. The initial guesses for the state variables are trivially chosen as the constant initial condition and the NN parameters are initialized randomly. No acceleration strategy or simulation is used for educated initial guesses. We set a maximum number of Ipopt iterations / epochs of $200$ and an optimality tolerance of $10^{-7}$.
	
	\subsubsection{Results} \label{sec:vdp_results}
	In almost all cases, the optimization terminates prematurely, since the optimality tolerance is fulfilled and thus, a local optimum was found. The corresponding training times are displayed in Table \ref{tab:timings_Vdp}. Note that, because the high noise ($\sigma = 0.5$) leads to larger objective values, the regularization $\lambda$ is increased in this case. Nevertheless, the optimization with a total of $2500$ discrete nodes is very rapid and terminates in several seconds from an extremely poor initial guess, while in some instances runs settle in poor local optima. Clearly, more reasonable initial guesses and sophisticated initialization strategies will further enhance these times and stability.
	\begin{table}[h]
		\centering
		\begin{tabular}{@{}ccrrrr@{}}
			\toprule
			$\sigma$ & $\lambda$ & \textbf{Total} & \textbf{Ipopt} & \textbf{Callbacks} & \textbf{\#Epochs} \\
			\midrule
			$0$  & $10^{-4}$ & 8.17 & 5.95 & 2.22 & 80 \\
			$0.1$  & $10^{-4}$ & 7.69  & 5.60 & 2.09 & 76\\
			$0.5$  &$10^{-3}$ & 13.49 & 9.79 & 3.70 & 134 \\
			\bottomrule
		\end{tabular}
		\vspace{10pt}
		\caption{Training Times (in seconds) and Number of Ipopt Iterations / Epochs for the Training the Van-der-Pol Oscillator NNs}
		\label{tab:timings_Vdp}
	\end{table}
	
\begin{figure}[h]
		\centering
		\includegraphics[width=0.6\textwidth]{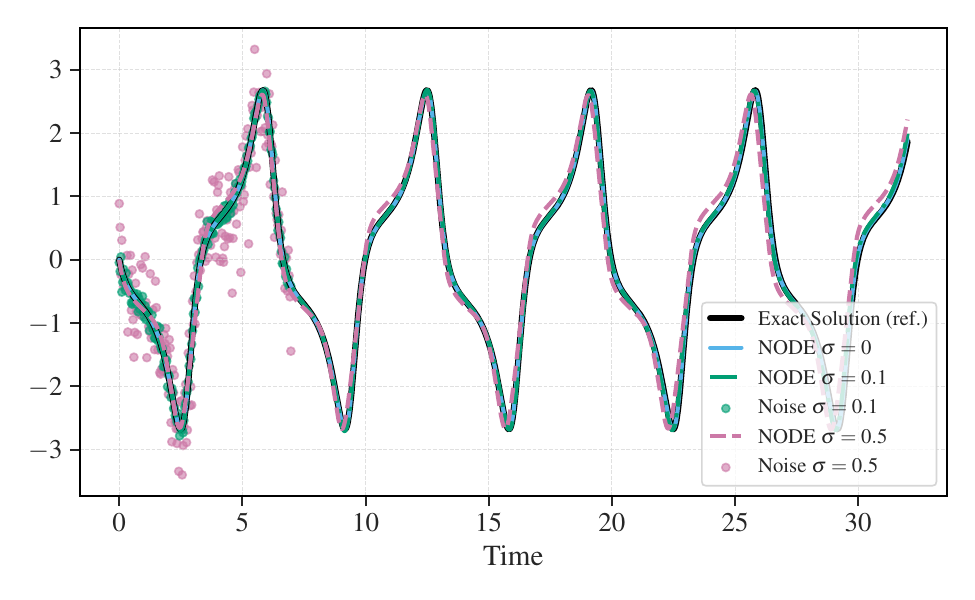}
		\caption{Simulation Results of the Neural and Reference Models as well as the Data for $y(t)$}
		\label{fig:vdp_sim_noisy}
	\end{figure}

	Figure \ref{fig:vdp_sim_noisy} presents the simulation results and training data for various levels of noise. It is evident that for no or low noise, the solution obtained is virtually indistinguishable from the reference, which is quite impressive. Even with high noise ($\sigma = 0.5$), the Neural ODE remains close to the true solution, showing significantly better performance compared to the results reported in \cite{Roesch2021}, where NODE and reference do not align for $\sigma \geq 0.2$. While smaller neural networks are employed here, these compact models still demonstrate exemplary performance and fast training, highlighting the effectiveness of the approach under severe noise.
	
	To verify the robustness of the training procedure, a comprehensive sensitivity analysis, consisting of 100 training runs for each noise level, was conducted. The results, detailed in the Appendix and Figure \ref{fig:vdp_combined}, demonstrate that while all runs converge perfectly for the no-noise case ($\sigma=0$) and approximately 95\% show excellent agreement under low noise ($\sigma=0.1$), as expected the robustness diminishes with high noise ($\sigma=0.5$). In these cases, several runs converge to poor local optima or fail to converge, leading to solutions with noticeable period and amplitude mismatches or an outright collapse of the trajectory.
		
	To further illustrate the obtained NODEs, we compare the learned and true vector fields of the ODE
	\begin{equation}
		\begin{bmatrix} x \\ y \end{bmatrix} \mapsto
		\begin{bmatrix} NN_{\vv{p}_x}^{x}(x, y) \\[0.2em]
			NN_{\vv{p}_y}^{y}(x, y) \end{bmatrix}, \quad
		\begin{bmatrix} x \\ y \end{bmatrix} \mapsto
		\begin{bmatrix} y \\
			\mu y \left(1 - x^2\right) - x \end{bmatrix}.
	\end{equation}
	
\begin{figure}[H]
		\centering
		\begin{minipage}{0.48\textwidth}
			\centering
			\includegraphics[width=1\textwidth]{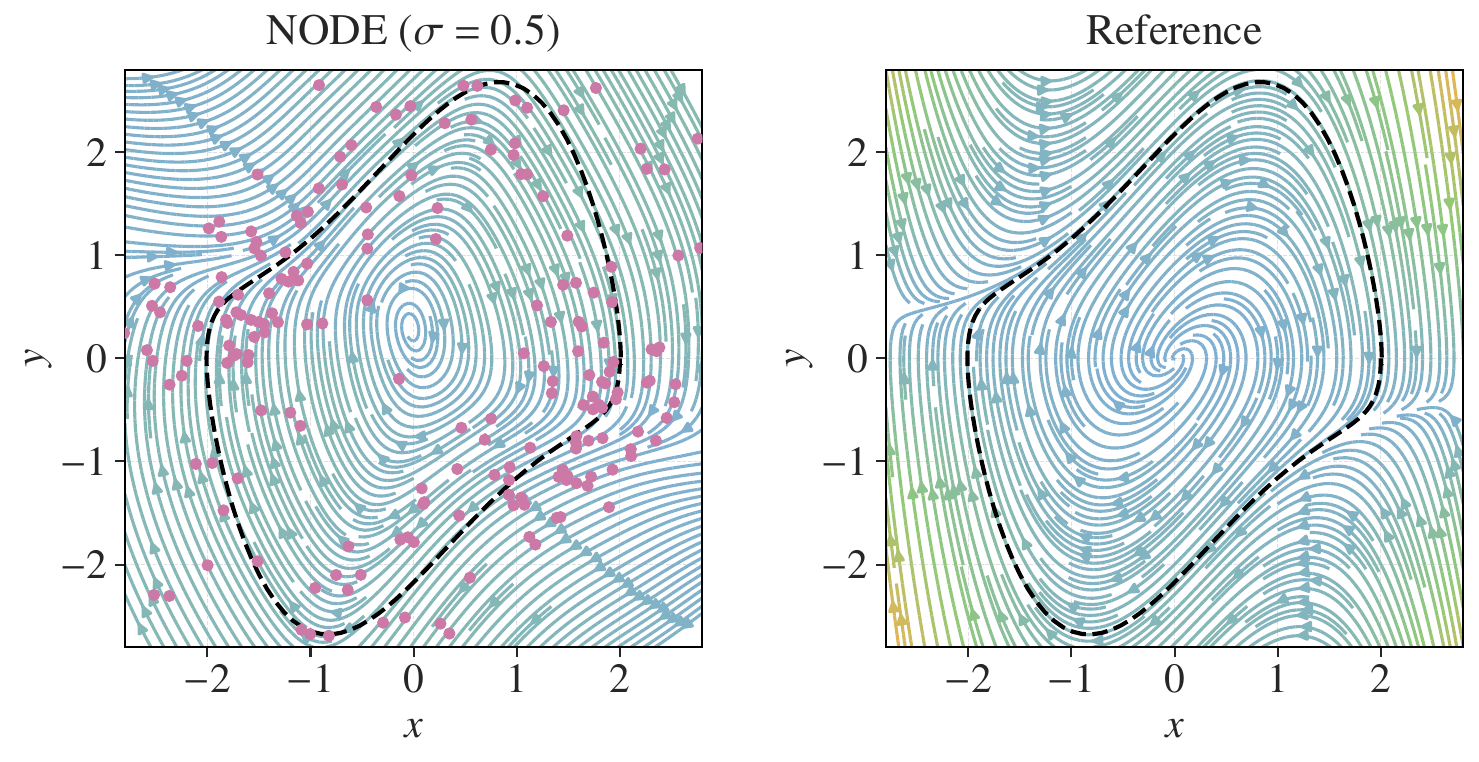}
			\vspace{-10.5pt}
			\caption{Neural ($\sigma = 0.5$) and Reference Vector Fields of the ODE and the Exact Van-der-Pol Trajectory (dashed)}
			\label{fig:vdp_vector_fields}
		\end{minipage}
		\hfill
		\begin{minipage}{0.48\textwidth}
			\vspace{2.5pt}
			\includegraphics[width=1\textwidth]{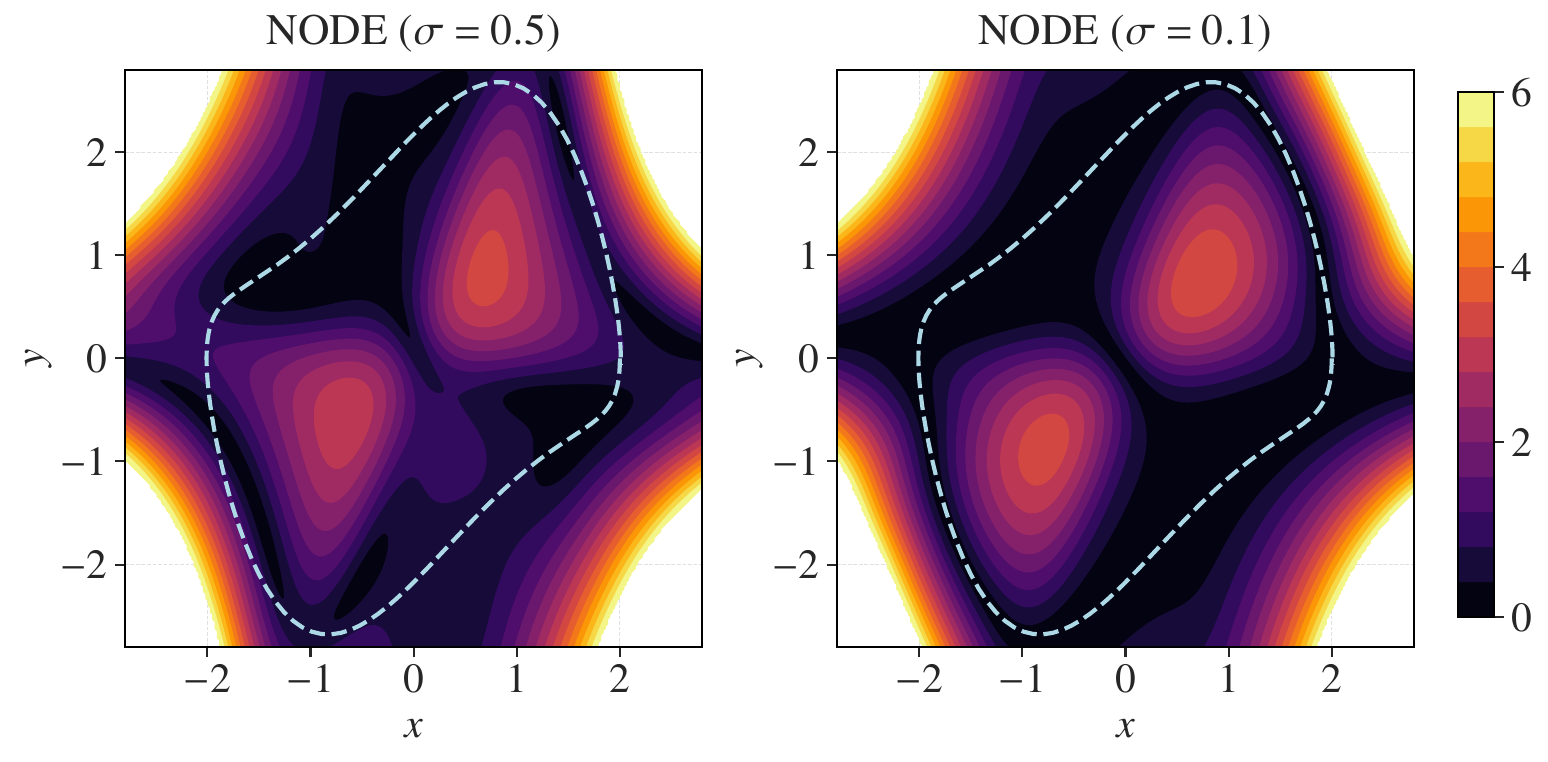}
			\vspace{-7.7pt}
			\caption{Scalar Fields Representing the $2$-Norm Error Between the Neural ($\sigma = 0.5$, $\sigma = 0.1$) and Reference Vector Fields (Values $> 6$ are white)}
			\label{fig:vdp_scalar_err}
		\end{minipage}
	\end{figure}

	In Figure \ref{fig:vdp_vector_fields}, we show the high-noise NODE, its training data, and the true vector field. Despite the heavily scattered observable states, the NODE still manages to recover a vector field that aligns well with the true dynamics in the vicinity of the solution trajectory.
	
	Further comparison is given in Figure \ref{fig:vdp_scalar_err}, where the scalar fields visualize the $2$-norm error between the neural and reference vector fields. The low-noise NODE ($\sigma = 0.1$) yields way smaller error values along the trajectory, but even the high-noise model produces a fairly accurate vector field in regions close to available training data. As expected, generalization outside this domain remains limited, but within the training region, the results demonstrate strong consistency.
	
	In addition to the local collocation approach with $500$ intervals, we also reproduce the results in \cite{Shapovalova2025} using a global, \textit{spectral} collocation method. We employ a single interval with $70$ fLGR nodes, corresponding to Radau IIA of order $139$. This high-order global formulation yields an approximation of equal quality to the figures above, even under high noise ($\sigma = 0.5$). Furthermore, the optimization terminates in outstandingly fast time, i.e. under $1.2$ seconds. This demonstrates the remarkable efficiency and accuracy of spectral methods for such smooth problems.
	
	\section{Future Work} \label{sec:modelica}
	Although OpenModelica currently includes an optimization runtime implementing direct collocation \cite{Ruge2014Collocation}, it remains limited to basic features: it supports only $1$- or $3$-step Radau IIA collocation, does not allow parameter optimization, lacks analytic Hessians, and does not perform parallel callbacks. To overcome these limitations, work is underway to embed an extended version of \textit{libgdopt} into OpenModelica. This integration combines recent developments from GDOPT with the existing strengths of OpenModelica, particularly its native ability to handle DAEs.
	The extended framework enables expressive Modelica-based modeling, native support of neural components via the NeuralNetwork Modelica library and incorporates recent advancements in mesh refinement for optimal control problems \cite{Langenkamp2024}.
	
	\subsection{NeuralNetwork Modelica Library}
	The NeuralNetwork library, originally developed in \cite{Codeca2006}, is an
	open-source Modelica library\footnote{\url{https://github.com/AMIT-HSBI/NeuralNetwork}} modeling ML
	architectures with pure Modelica.
	Neural components can be constructed by connecting dense feedforward layers of
	arbitrary size with layers for PCA, standardizing, or scaling.
	These blocks contain all equations and parameters as pure Modelica code, which
	makes seamless integration of neural components into existing Modelica models
	straightforward.
	Together with this library, OpenModelica will enable modeling and training of
	PeN-ODEs within a single development environment.
	
	\subsection{Workflow}
\begin{figure*}[h]
		\centering
		\includegraphics[width=0.85\textwidth]{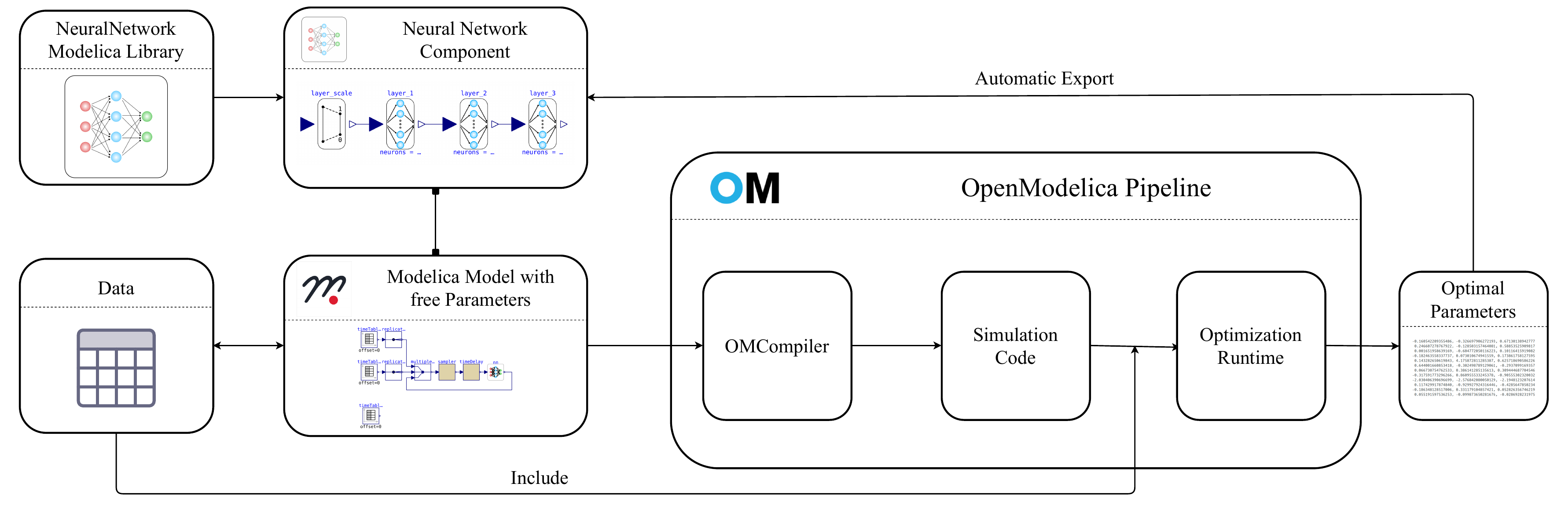}
		\caption{OpenModelica Workflow for PeN-ODE Training (under development)}
		\label{fig:mo_workflow}
	\end{figure*}

	The workflow presented in Figure \ref{fig:mo_workflow} illustrates the intended process for native Modelica-based modeling and training of PeN-ODEs. While some components of the workflow are operational, the full integration is still under active development. Users model physical systems and NN components with free parameters directly in Modelica and provide corresponding data. Both the model and data are processed by the standard OpenModelica pipeline, while the OpenModelica Compiler (OMC) generates fast C code.
	The new backend of OMC introduces efficient array-size-independent symbolic manipulation \cite{Abdelhak2023}. Ongoing work further targets resizability of components after compilation \cite{Abdelhak2025Workshop}, offering significant benefits for array-based components such as NNs.
	
	Furthermore, current work focuses on leveraging the generated simulation code to enable fast callback functions for a new optimization runtime. After training, the optimal parameters are directly inserted into a NN block, enabling immediate simulations by swapping connectors.
	
	This workflow removes the need for export and import steps for neural components and models, e.g. in the form of a Functional Mock-up Unit (FMU). It also eliminates reliance on external training routines in Julia or Python and avoids external C functions in the model, since the neural components are pure Modelica blocks. This integrated workflow unifies modeling, training, and simulation within a single toolchain, enabling free and accessible optimizations.
	
	\subsection{Neural DAEs}
	One of the main advantages of the integration into OpenModelica is the native ability to handle DAEs. Modelica compilers such as OpenModelica systematically apply index reduction and block-lower triangular (BLT) transformations to restructure DAEs into semi-explicit ODE form with index 1 \cite{Ruge2014Collocation, Akesson2012DAE}. As the simulation code already resolves algebraic variables during evaluations, no additional handling, e.g. inclusion of algebraic variables in the NLP, is needed on the optimization side. This allows the new training workflow to seamlessly extend from ODEs to DAEs, far surpassing the current range of applications.
	
	\section{Conclusion} \label{sec:conclusion}
	This paper proposes a formulation of PeN-ODE training as a collocation-based NLP, simultaneously optimizing states and NN parameters. The approach overcomes key limitations of ODE solver-based training in terms of order, stability, accuracy, and allowable step size. The NLP uses high order quadrature for the NN loss, potentially preserving the accuracy of the discretization. We demonstrate that known physical behavior can be trivially enforced. 
	
	We provide an open-source parallelized extension to GDOPT and on two example problems
	demonstrate exemplary accuracy, training times, and generalization with smaller NNs compared to other training techniques, even under significant noise. Furthermore, we show that the approach allows for efficient optimization of other parameter dependent surrogates.
	
	Key limitations and challenges of the proposed method, including grid selection and general initialization strategies to increase stability, as well as training with larger datasets and networks are identified. Addressing and evaluating these issues in future work is essential to support broader applicability. To enable accessible training of Neural DAEs, without relying on external tools, work is underway to implement the method in OpenModelica.
	
	\section*{Acknowledgements}
	This work was conducted as part of the OpenSCALING project (Grant No. 01IS23062E) at the University of Applied Sciences and Arts Bielefeld, in collaboration with Linköping University.
	The authors would like to express their sincere appreciation to both the OpenSCALING project and the Open Source Modelica Consortium (OSMC) for their support,
	collaboration, and shared commitment to advancing open-source modeling and simulation technologies.
	
	%\bibliographystyle{unsrtnat}
	%\bibliography{references}

	\newpage
	\section*{Appendix: Sensitivity Analysis and Robustness of VdP Neural ODEs}
	\begin{figure}[H]
		\centering
		\begin{minipage}[b]{0.33\textwidth}
			\includegraphics[width=\textwidth]{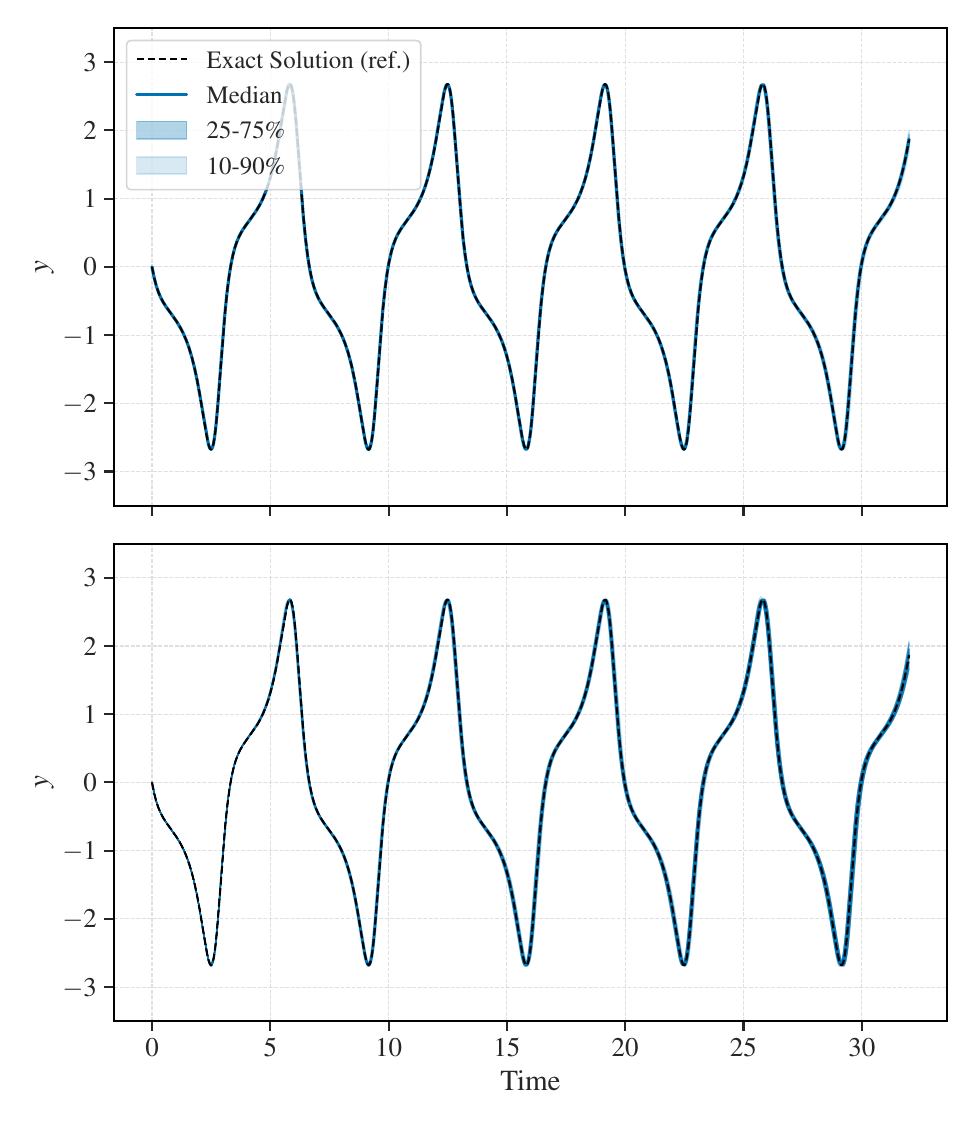}
			\subcaption{No noise: $\sigma=0$}
		\end{minipage}
		\hfill
		\begin{minipage}[b]{0.33\textwidth}
			\includegraphics[width=\textwidth]{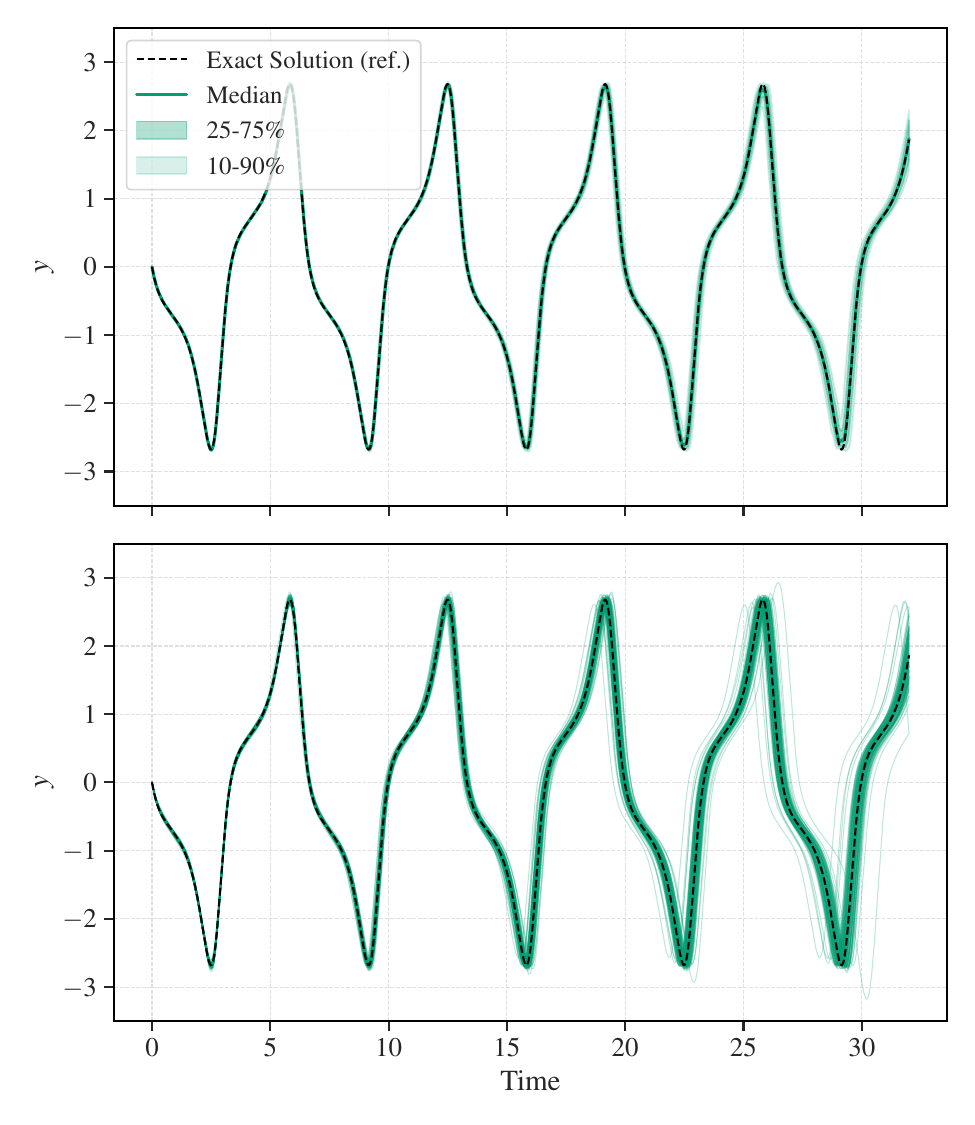}
			\subcaption{Low noise: $\sigma=0.1$}
		\end{minipage}
		\hfill
		\begin{minipage}[b]{0.33\textwidth}
			\includegraphics[width=\textwidth]{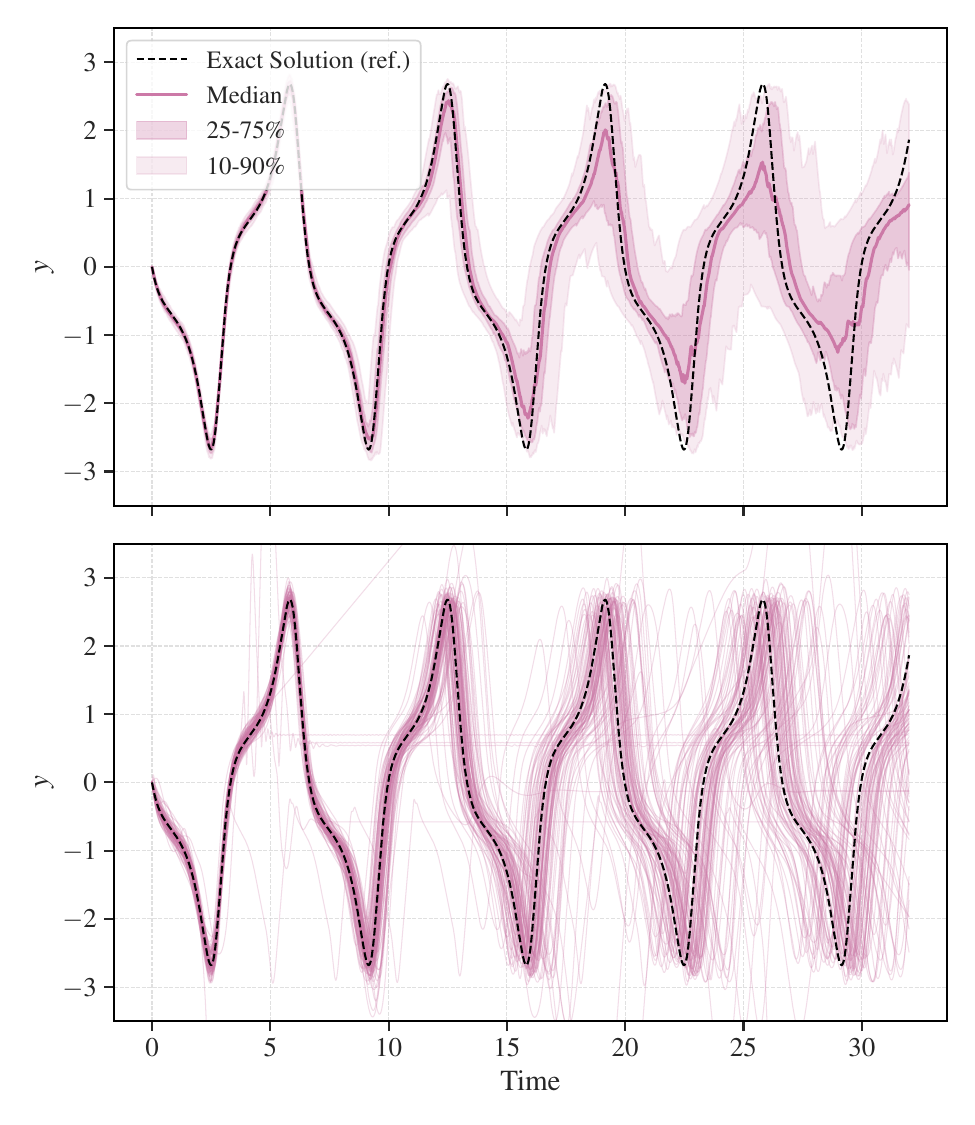}
			\subcaption{High noise: $\sigma=0.5$}
		\end{minipage}
		\caption{
			Sensitivity analysis of the learned Van-der-Pol (VdP) Neural ODEs to different levels of Gaussian noise. For each noise level, 100 training sessions were performed with random initial guesses for the neural network parameters and a random seed for the data perturbation. The plots show the simulation results for the observable state y(t) over an extended time horizon of 32 seconds. Each subplot consists of two panels: The top panel visualizes the reference solution (dashed black line), the median of the 100 learned solutions, and the 25-75\% and 10-90\% percentile bands. The bottom panel shows all 100 individual learned trajectories (faint lines).
			\textbf{(a) No noise ($\sigma=0$):} The method demonstrates full robustness and perfect convergence, with all 100 solutions converging to the exact reference solution.
			\textbf{(b) Low noise ($\sigma=0.1$):} The method remains highly robust. A vast majority of the solutions (approximately 95\%) form a tight band around the reference, showing excellent agreement.
			\textbf{(c) High noise ($\sigma=0.5$):} As expected under severe noise, the robustness is reduced. While many solutions remain relatively close to the reference, some diverge significantly, indicating a failure to find a good local optimum during training. The solutions that do converge often show a period mismatch or diverge after a few periods, but still capture the general oscillatory behavior. The combined effects of reduced solution quality and an inaccurate period are reflected in the median and percentile bands, which exhibit a noticeable deviation in amplitude and phase after three periods compared to the true trajectory.
		}
		\label{fig:vdp_combined}
	\end{figure}
\end{document}